\pdfoutput=1

\documentclass[11pt]{article}

\usepackage{emnlp2021}

\usepackage{times}
\usepackage{latexsym}

\usepackage[T1]{fontenc}

\usepackage[utf8]{inputenc}
\usepackage{soul}

\usepackage{microtype}

\usepackage{tikz,pgfplots}

\title{Is ``moby dick'' a Whale or a Bird? \\ Named Entities and Terminology in Speech Translation}

\author{Marco Gaido\textsuperscript{1,2}, Susana Rodríguez\textsuperscript{3}, Matteo Negri\textsuperscript{1}, Luisa Bentivogli\textsuperscript{1} and Marco Turchi\textsuperscript{1} \\
  \textsuperscript{1}Fondazione Bruno Kessler \\
  \textsuperscript{2}University of Trento \\
  \textsuperscript{3}Independent Researcher \\
  \texttt{\{mgaido,negri,bentivo,turchi\}@fbk.eu} \\}

\begin{document}
\maketitle
\begin{abstract}

Automatic translation systems are known to struggle with rare words. Among these, named entities (NEs) and  domain-specific terms are crucial, since errors in their translation can lead to severe meaning distortions. Despite their importance, previous speech translation (ST) studies have neglected  them, also due to the dearth of publicly available resources tailored to their specific evaluation. To fill this gap, we \textit{i)} present the first systematic analysis of the behavior of state-of-the-art ST systems in translating NEs and terminology, and \textit{ii)} release NEuRoparl-ST, a novel benchmark built from European Parliament speeches annotated with NEs and terminology. Our experiments on the three language directions covered by our benchmark (en$\rightarrow$es/fr/it) show that ST systems correctly translate 75–80\% of terms and 65–70\% of NEs, with very low performance (37–40\%) on person names.
\end{abstract}

\section{Introduction}
The translation of rare words is one of the main challenges for neural machine translation (NMT) models
\cite{sennrich-etal-2016-neural,koehn-knowles-2017-six}.
Among rare words, named entities (NEs) and  terminology are particularly critical: not only are they important to understand the meaning of a sentence \cite{li-etal-2013-name}, but they are also difficult to handle due to the small number of valid translation options. 
While common words can be rendered in the target language with synonyms or paraphrases, NEs and terminology offer less expressive freedom, which is typically limited to one valid option.
Under these conditions, translation errors often result in blatant (meaningless, hilarious, or even offensive) errors, which jeopardize users' trust in the translation system. One example is ``moby dick''  (in lower case, as in the typical output of a speech recognition system): Google Translate\footnote{Accessed on the 27th April 2021.} returns  \textit{mazikó poulí} (massive bird) for Greek, while the translation contains profanities
for other languages like Hungarian or Italian.

Previous works in NMT tried to 
mitigate the problem by: \textit{i)} integrating knowledge graphs (KG) into the models \cite{lu-2019-exploiting,Moussallem-2019-Utilizing,zhao-etal-2020-knowledge-graph,zhao-2020-knowledge-graph,Ahmadnia_Dorr_Kordjamshidi_2020}, \textit{ii)} exploiting dedicated modules for NE translation 
\cite{Yan-2019-impact}, or \textit{iii)}
adding  NE tags to the source  \cite{ugawa-etal-2018-neural,Zhou-2020-incorporating}.
For terminology, the 
so far proposed dictionary-based approaches 
\cite{hokamp-liu-2017-lexically,chatterjee-etal-2017-guiding,hasler-etal-2018-neural,dinu-etal-2019-training,Song_Wang_Yu_Zhang_Huang_Luo_Duan_Zhang_2020,dougal-lonsdale-2020-improving}
share  the idea of enriching  the source sentence with 
terms' translations
found in a dictionary  (e.g. ``this is a term''$\rightarrow$``this is a \#term\#\textit{término}\#''). 

The problem is even more challenging in automatic speech recognition (ASR) and speech-to-text translation (ST), where a lookup into KGs or dictionaries is not feasible due to the different modality of the input (an audio signal rather than text). As regards NEs, the few 
existing studies \cite{ghannay2018endtoend,caubriere-etal-2020-named} are all limited to 
ASR, for which two benchmarks are available \cite{galibert-etal-2014-etape,Yadav2020}, 
while  suitable benchmarks do not even exist for ST.
The situation is similar for terminology: 
few  annotated test sets exist for MT  \cite{,dinu-etal-2019-training,scansani-etal-2019-magmatic,bergmanis-pinnis-2021-facilitating}, 
but none for ST, which so far has remained unexplored.

In light of the above, the contribution of 
this work is twofold: \textbf{(1)} we present  the  first  investigation on the behavior of state-of-the-art ST systems in translating NEs and terms, discussing their weaknesses and providing baseline results for future comparisons; \textbf{(2)} we release the annotated data that made our
study possible. 
Our test set -- NEuRoparl-ST -- is derived from Europarl-ST \cite{jairsan2020a} and covers three language pairs: 
en$\rightarrow$es/fr/it.
It relies on the Europarl-ST (\textit{audio}, \textit{transcript}, \textit{translation}) triplets and enriches their textual portions with NE and terminology annotation.
Besides being  the first benchmark of this type for ST, it can also be  used  for the evaluation of NE/terminology recognition (ASR) and translation (MT). 
The dataset
is available at: \url{ict.fbk.eu/neuroparl-st/}.

\section{Speech Translation Models}
\label{sec:models}

Our goal is to assess the capability of state-of-the-art ST systems to properly translate NEs and terminology present in an utterance. To this aim, we compare instances of the two main approaches. One is the traditional
\textit{cascade} approach \cite{StentifordSteer88,Waibel1991b},
which consists of a pipeline  where  an ASR model 
produces a transcript
of the input audio
and an MT model generates its translation.
The other is the so-called \textit{direct} approach \cite{berard_2016,weiss2017sequence}, which relies on a single neural network that maps the audio into target language text bypassing any intermediate symbolic representation.
The two approaches have inherent strengths and weaknesses \cite{sperber-paulik-2020-speech}. 
Cascade solutions can exploit sizeable datasets
for the 
ASR and MT sub-components, but rely on a complex architecture prone to error propagation.
Direct models suffer from the paucity of training data, but avoid error propagation and can take advantage of unmediated access to audio information (e.g.  prosody) during the translation phase.
In recent years,
after a long dominance of the cascade paradigm, the initially huge performance gap between the two approaches has gradually closed \cite{ansari-etal-2020-findings}.

Our \textbf{cascade} system integrates competitive Transformer-based \cite{transformer} ASR and MT  components
built from large training corpora. Specifically, the ASR model is trained on LibriSpeech \cite{7178964}, TEDLIUM v3 \cite{DBLP:conf/specom/HernandezNGTE18} and Mozilla Common Voice,\footnote{\url{http://commonvoice.mozilla.org/en/}} together with  (\textit{utterance}, \textit{transcript}) pairs extracted from  
 MuST-C \cite{MuST-Cjournal}, Europarl-ST \cite{jairsan2020a}, and CoVoST 2 \cite{wang-etal-2020-covost} ST corpora. ASR outputs are post-processed to add true-casing and punctuation. 
 The MT model is trained on data collected from the OPUS repository,\footnote{\url{http://opus.nlpl.eu}} amounting to about 19M, 28M, and 45M parallel sentence pairs respectively for en-es, en-fr, and en-it.

Our \textbf{direct} model has the same Transformer-based architecture of the ASR component used in the cascade system. It exploits data augmentation  and knowledge transfer techniques successfully applied by participants in the IWSLT-2020 evaluation campaign \cite{ansari-etal-2020-findings,potapczyk-przybysz-2020-srpols,gaido-etal-2020-end} and it is trained on MuST-C, Europarl-ST and 
synthetic data ($\sim$1.5M pairs for each language direction).

Systems' performance is shown in Table \ref{tab:st_and_mt} and discussed in Section \ref{sec:results}.
Complete details about their implementation and 
training procedures are provided in the 
Appendix.
All the related code is available at \url{https://github.com/mgaido91/FBK-fairseq-ST/tree/emnlp2021}.

\section{Evaluation Data: NEuRoparl-ST}
\label{sec:benchmark}

\begin{table*}[htb]
\small
\center

\begin{tabular}{l|c|c|c|c|c|c}
 &
  \multicolumn{2}{c|}{\textbf{en-es}} &
  \multicolumn{2}{c|}{\textbf{en-fr}} &
  \multicolumn{2}{c}{\textbf{en-it}} \\
 &
  \textbf{en} &
  \textbf{es} &
  \textbf{en} &
  \textbf{fr} &
  \textbf{en} &
  \textbf{it} \\
  \hline
NEs &
  1,637 (2,703) &
  1,638 (3,003) &
  1,578 (2,604) &
  1,562 (2,949) &
  1,523 (2,497) &
  1,466 (2,649) \\
TERMS &
  2,571 (3,174) &
  2,662 (3,294) &
  2,797 (3,502) &
  2,947 (3,659) &
  2,166 (2,669) &
  2,202 (2,645)
\\
\hline
Num. of sentences &
  \multicolumn{2}{c|}{1,267} &
  \multicolumn{2}{c|}{1,214} &
  \multicolumn{2}{c}{1,130}
\end{tabular}
\caption{\label{tab:stats}Total number of named entities and terms annotated in the test sets (and corresponding number of tokens).}
\end{table*}

To the best of our knowledge, freely available NE/term-labelled ST benchmarks suitable for our analysis do not exist. 
The required resource
should contain  \textit{i)} the audio corresponding to an utterance, \textit{ii)} its transcript, \textit{iii)} its translation in 
multiple
target languages (three in our case), and \textit{iv)} NE/term annotation in both transcripts and target texts. Currently
available MT, ST, ASR, NE and terminology datasets lack at least one of these key components. For example, most MT corpora (e.g. Europarl) lack both the audio sources and NE/terminology annotations. The very few available MT corpora annotated with NE/terminology still lack the audio portion, and extending them to ST would require generating synthetic audio, which is known to be problematic for models’ performance. 
For these reasons, we
preferred to create a 
the 
en$\rightarrow$es/fr/it
transcripts and translations 
of the Europarl-ST test sets, which are mainly derived
from the same original speeches. 
The result is 
a multilingual benchmark featuring very high content overlap, thus enabling cross-lingual 
comparisons.

\noindent
\textbf{NE annotation.}
We used the 18  tags  and the annotation scheme defined by the guidelines (``OntoNotes Named Entity Guidelines - Version 14.0'') used to annotate the OntoNotes5 corpus \cite{ontonotes}.
The annotation
was carried out manually by a professional interpreter with a multi-year experience in translating from English, French and Italian into Spanish the verbatim reports of the 
European Parliament plenary meetings.
This guarantees the high level of language knowledge and domain
expertise required to ensure maximum quality and precision.
To ease the task, the annotator was provided with transcripts and translations automatically pre-annotated with  the BERT-based NER model\footnote{\url{http://docs.deeppavlov.ai/en/master/features/models/ner.html}} available in
DeepPavlov \cite{burtsev-etal-2018-deeppavlov}.
Human annotation was then conducted in parallel on the three test sets by labelling, for each audio segment, the English transcript and the three corresponding translations. 
To check annotations' reliability, all the English transcripts were also independently labelled
by a second annotator with a background in linguistics and  excellent English knowledge. Inter-annotator agreement was calculated in terms of \textit{complete} agreement, i.e. the exact match of the whole NE in the two annotations.
The resulting Dice coefficient\footnote{Note that Dice coefficient has the same value of the F1 measure computed considering either annotator as reference.} \citep{dice}
%
amounts to 93.87\% and can be considered highly satisfactory. 
For the subset of NEs for which complete agreement  was  found (1,409 in total), we also  computed
the agreement on  labels' assignment with the {\em kappa coefficient} (in Scott's  $\pi$ formulation) \cite{10.1086/266577,Artstein:2008:IAC:1479202.1479206}.
The resulting value is 0.94,
which  corresponds to ``almost perfect'' agreement according to its standard interpretation \cite{Landis77}.

\noindent
\textbf{Terminology annotation.}
Similar to \cite{dinu-etal-2019-training}, terminology was automatically extracted
by exploiting the IATE 
term base.\footnote{\url{http://iate.europa.eu}} Each entry in IATE has an identifier and a language code. Entries with the same identifier and different language codes represent the translations of a term in the corresponding languages.
To annotate our parallel texts, we first removed stop-words and lemmatized the remaining words 
and IATE 
entries.\footnote{Preprocessing made with spaCy: \url{http://spacy.io/}} 
Then, for each parallel sentence, we marked as terms only those words in the source and the target side that were present in IATE with the same identifier.
This source/target match is essential to avoid the annotation of words that are used with a generic, common 
meaning but, being polysemic,
can be technical terms in different contexts (e.g. the word ``board'' can refer to a tool or to a committee). Checking the presence of the corresponding translation in the target language disambiguates these cases, leading to a more accurate annotation.

NE and term annotations were merged into a single test set using BIO \cite{ramshaw-marcus-1995-text} as span labeling format. Had a word been tagged both as term and
NE, the latter was chosen favoring the more reliable manual annotation.
Table \ref{tab:stats} presents the total number of NEs and terms for the three language pairs, together with their corresponding number of tokens.\footnote{The statistics
for each NE type are given in the Appendix.}
These numbers differ between source and target texts and across pairs
due to
the peculiarities of the Europarl-ST data. Specifically, \textit{i)} sometimes translations are not literal and NEs are
omitted in the translation (e.g. when a NE is repeated in the source, one of the occurrences may be replaced by a pronoun in the target text), \textit{ii)} the professional interpreters and translators ``localize'' the target translations, i.e. adapt them to the target culture (e.g. while the English source simply contains the name and surname of mentioned European Parliament members,  in Italian the first name is omitted and the surname is preceded by ``onoverole'' - honorable), and \textit{iii)} the number of words a NE is made of can vary across languages (e.g ``European Timeshare Owners Organisation'' becomes ``Organización Europea de Socios de Tiempo Compartido'' in Spanish).

\begin{table*}[htb]
\small
\center
\begin{tabular}{l|cccc|cccc|cccc}
 &
  \multicolumn{4}{c|}{\textbf{en-es}} &
  \multicolumn{4}{c|}{\textbf{en-fr}} &
  \multicolumn{4}{c}{\textbf{en-it}} \\
  &WER
 &
  BLEU &
  NE &
  Term &
   WER
   & BLEU &
  NE &
  Term &
   WER &
  BLEU &
  NE &
  Term 
  \\
  \hline
  ASR & 12.6 & -- & 84.6 & 92.6 & 12.7 & -- & 84.5 & 92.1 & 12.6 & -- &  84.3 & 92.4 \\
  MT &   -- & 48.8    & 83.5 &  88.8   & -- & 36.2 & 78.8 & 85.7  & -- & 33.8 & 80.2  & 86.9  \\
  \hline
  Cascade &  -- & 37.6 & 70.9 & 82.5  & -- & 28.3 & 66.7 & 80.4  & -- & 26.5 & 66.9  & 80.4 \\
  \hline
  Direct & -- & 37.7 & 71.4 & 79.2 & -- & 30.1 & 67.3 &  77.7  & -- & 26.0  & 67.3 & 76.3
 \\

\hline
\end{tabular}
\caption{\label{tab:st_and_mt}
WER/BLEU and NE/term case-insensitive accuracy for ASR, MT and ST (cascade and direct) models.}
\end{table*}

\section{Results}
\label{sec:results}
We use our benchmark to measure systems' ability to handle NEs and terminology. 
Besides comparing the two ST models described in Section \ref{sec:models}, we extend our evaluation to the
ASR and MT sub-components (the latter being fed with human transcripts) of the cascade system. 
As shown in Table \ref{tab:st_and_mt}, all  models are evaluated in terms of overall output quality 
and accuracy in rendering the two categories of rare words subject of our study.
Transcription and translation quality are respectively measured with  WER and SacreBLEU\footnote{\texttt{BLEU+c.mixed+\#.1+s.exp+tok.13a+v.1.5.0}} \cite{post-2018-call}. 
Similarly to the Named Entity Weak Accuracy proposed in \cite{hermjakob-etal-2008-name}, we compute  NE/term
accuracy\footnote{Scores have been computed with the script available at: \url{https://github.com/mgaido91/FBK-fairseq-ST/blob/emnlp2021/scripts/eval/ne_terms_accuracy.py}} as the ratio of entities that are present in the systems' output in the  correct 
form.
We present case-insensitive accuracy scores to fairly compare the different models, as the ASR produces lowercase text.

For the sake of completeness, case-sensitive NE/term accuracy is also given in Table \ref{tab:st_and_mt_cs} for ST and MT models (we do not include ASR since it generates lowercase text).
Comparing these results with those reported in Table \ref{tab:st_and_mt}, for all language pairs we see that the drop in NEs accuracy with respect to case-insensitive scores is higher for the cascade model -- around 5 points -- than for the direct one -- around 2 points (e.g. for en-es, from 70.9 to 65.8 for the cascade model and from 71.4 to 69.4 for the direct model).
We posit the reason is the propagation of errors in the module in charge to restore casing on the ASR output in the cascade architecture.

\begin{table}[htb]
\small
\center
\begin{tabular}{l|cc|cc|cc}
 &
  \multicolumn{2}{c|}{\textbf{en-es}} &
  \multicolumn{2}{c|}{\textbf{en-fr}} &
  \multicolumn{2}{c}{\textbf{en-it}} \\
 &
  NE &
  Term &
  NE &
  Term &
  NE &
  Term 
  \\
  \hline
  MT & 81.0 &  88.0 & 75.5 & 85.3  & 77.5  & 86.2  \\
  Cascade & 65.8 & 81.6  & 61.3 & 79.9  & 62.6  & 79.5 \\
  Direct & 69.4 & 78.7 & 65.9 &  77.3  & 65.1 & 75.9
 \\

\hline
\end{tabular}
\caption{\label{tab:st_and_mt_cs}
Case sensitive accuracy scores of MT and ST (cascade and direct) models on en$\rightarrow$es/fr/it.}
\end{table}

\subsection{ASR and MT results}
The WER of the ASR 
is similar across the three language directions.
This is not surprising because the three test sets differ
only in very few debates. In terms of 
accuracy, it is evident that transcribing NEs is more difficult than transcribing terms (84.5 vs 92.4 on average).
Besides lower frequency, the higher difficulty to transcribe NEs can be ascribed to the variety of different pronunciations by non-native speakers
(in particular for person, product and organization names).
Concerning
the MT performance, the BLEU differences between language directions (en-es $\gg$ en-fr $>$ en-it) reflect the results reported in the Europarl-ST paper \cite{jairsan2020a}. The main reason is that the 
translations are less literal for some language directions. For instance, the French references are 20\% longer than the  human source transcripts. Analyzing NE and term translation quality, we notice that NEs are, again,  
harder to handle compared to terminology (average accuracy: 80.8 vs 87.1). 
It is worth to notice that accuracy 
does not strictly depend on
translation quality. 
For instance,
en-fr has a 
higher
translation quality than en-it (+2.4 BLEU points), but 
NE and term accuracy scores are lower.

\subsection{ST results}
Unsurprisingly, when it comes to combining transcription and translation in a single task, performance
decreases significantly. 
In particular, the results of the cascade model are a direct consequence of cumulative  ASR and MT errors.
As such, like for its sub-components, NEs are harder to handle than terms. Compared to MT results computed on manual transcripts, 
we see large drops in all languages on both translation quality (-13.2 BLEU on average) and NE/term accuracy (-12.8/-6.0).

Comparing cascade and direct models, the BLEU scores are on par for en-es and en-it (differences are not statistically significant\footnote{Computed with bootstrap resampling \cite{koehn-2004-statistical}.}), while the direct one is significantly better for en-fr.
This is explained by the 
aforementioned
peculiarity of the French reference translations in Europarl-ST that,
unlike in common training corpora 
(Europarl included), 
are on average 20\%  longer than the source transcripts.
The MT model of the cascade, 
trained on
massive corpora
including Europarl,
tends to produce translations that are similar in length to the transcripts and
shorter than Europarl-ST references,
being thus penalized.
Having Europarl-ST 
among
its training corpora, 
the direct model produces outputs
more similar in length to the references,
resulting in a 2.8 BLEU gain. 

In terms of 
NE and term
translation quality, the trend is 
clear and coherent on all languages:
the cascade outperforms the direct on terminology (+3.5 on average), while the direct has an edge (+0.5) in handling NEs. 
The advantage of the cascade on terminology
can be explained by the higher reliability of its MT component in selecting domain-specific target words compared to the direct models built on much smaller ST training corpora.
One example is the English term ``plastic explosive'', which is correctly translated into Italian by the cascade (``esplosivo plastico''),  and wrongly by the direct (``esplosivo di plastica'' - En: ``explosive made of plastic'').
Concerning NEs, instead, the 
unmediated
access to the audio helps the direct to avoid both \textit{i)} error propagation (e.g. the NE ``Lamfalussy'' is correctly translated by the direct, while the MT component of the cascade
is not able to recover the wrong ASR output ``blunt Hallucy''), and \textit{ii)} 
the translation of NEs that are homographs of common nouns in the source language but should be copied \textit{as is} (e.g. the English surname ``Parish'' is translated into Italian as ``Parrocchia'' by the cascade, but correctly preserved  in the direct's output).

\begin {figure}[btp]
\begin{tikzpicture}[scale=0.9]
\begin{axis}[
symbolic x coords={PERSON,LOCATION},
    xtick=data,
	ylabel=Accuracy,
	legend style={at={(0.5,-0.3)},
	anchor=north,
	legend columns=-1},
	ybar=5pt,
	bar width=12pt,
	width=7cm,
	height=5cm,
	enlarge x limits=0.5,
]
\addplot 
	coordinates {(PERSON,0.9375) (LOCATION,0.8571)};
\addplot 
	coordinates {(PERSON,0.4063) (LOCATION,0.7912)};
\addplot 
	coordinates {(PERSON,0.3854) (LOCATION,0.8132)};
\addplot 
	coordinates {(PERSON,0.4478) (LOCATION,0.9339)};
\legend{MT,Cascade,Direct,ASR}
\end{axis}
\end{tikzpicture}
\caption{\label{fig:person_and_gpe} Accuracy scores on PERSON and LOCATION of MT, ASR and ST systems on en-es.}

\end{figure}
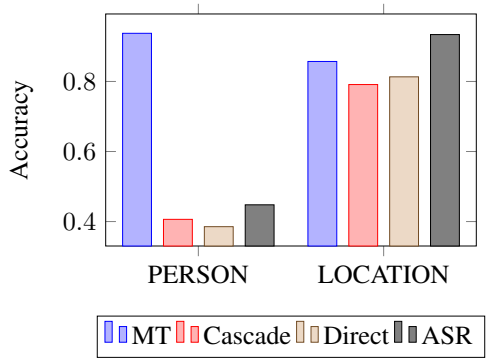

Looking at NE types (complete results in the 
Appendix), the two ST systems
are always close to each other, reflecting the global accuracy scores in Table \ref{tab:st_and_mt}. For both approaches, the differences across the NE types depend on their capability to \textit{recognize} entities in the audio and properly \textit{translate} them. Two types are paradigmatic 
(see Figure \ref{fig:person_and_gpe}). 
PERSON names (the worst category, with 37–40\% ST accuracy) are difficult to recognize
in the audio, as shown by the poor performance of ASR and both ST systems, while their translation from manual transcripts (MT) is trivial as it only requires copying them from the source. Conversely, ST and MT results are very close  on the more frequent and normally easier to pronounce LOCATION names, for which the problem lies more in translation than in recognition.

\section{Conclusions}

While previous ST research has 
focused on improving overall  systems' performance, little has been done to 
evaluate the existing paradigms in relation to well known specific problems in automatic translation at large. 
Translating
rare words is no exception, also due to the dearth of 
suitable labelled benchmarks.
To fill this gap, we focused on named entities and terminology, which combine the problems inherent to low frequency in the training data with the  difficulty of recognizing them in the audio and mapping their meaning into few valid options. We created NEuRoparl-ST, an annotated benchmark covering three language directions, and used it  for the first comparison of state-of-the-art cascade and direct ST systems on NE and term translation. Our results show that NEs, especially person names, are in general more difficult to handle than terminology.

\section*{Acknowledgement}

This work has been carried out as part of the project Smarter Interpreting (\url{https://kunveno.digital/}) financed by CDTI Neotec funds.

\bibliography{anthology,custom}
\bibliographystyle{acl_natbib}

\appendix

\section{Models and Trainings}

\subsection{Cascade ST Model}
\label{sec:cascade}

The \textbf{ASR} component of our cascade is a Transformer-based \cite{transformer} model consisting of 11 encoder layers, 4 decoder layers, 8 attention heads, 512 features for the attention layers and 2,048 hidden units in the feed-forward layers.  Its encoder has been adapted for processing speech by means of two initial 2D convolutional layers that reduce the input sequence length by a factor of 4. Also, the encoder self-attentions are biased using a logarithmic distance penalty that favors the local context \cite{di-gangi-etal-2019-enhancing}. Similar to \cite{gaido-etal-2020-end}, the model is trained with an additional Connectionist Temporal Classification (CTC) loss \cite{Graves2006ConnectionistTC}, which is added as a linear layer to the 8th encoder layer. As training data, we used LibriSpeech \cite{7178964}, TEDLIUM v3 \cite{DBLP:conf/specom/HernandezNGTE18} and Mozilla Common Voice,\footnote{\url{https://commonvoice.mozilla.org/en/datasets}} together with  (\textit{utterance}, \textit{transcript}) pairs extracted from  three ST corpora: MuST-C \cite{MuST-Cjournal}, Europarl-ST, and CoVoST 2 \cite{wang-etal-2020-covost}. We augment data with SpecAugment \cite{Park_2019} and, after lowercasing and punctuation removal, text is split into sub-words with 8,000 BPE~\cite{sennrich-etal-2016-neural} merge rules.
We set the dropout to 0.1. We optimize label smoothed cross entropy with smoothing factor 0.1 with Adam \cite{kigma-ba-2015-adam}. The learning rate is increased for 5,000 steps from 0.0003 up to 0.0005 and then decays with inverse square root policy.
Our mini-batches are composed of up to 12K tokens or 8 samples and we delay parameter updates for 8 mini-batches. We train on 8 GPU K80 (11GB RAM). 

Before feeding the MT with the ASR outputs, the transcripts are post-processed by an additional model to restore casing and punctuation. This model is a Transformer-based system
trained on data
from the OPUS repository, where the source text is lowercased 
and without punctuation and the target text is a normally formatted sentences.

The \textbf{MT} component is a Transformer model with 6 layers for both the encoder and the decoder, 16 attention heads, 1,024 features for the attention layers and 4,096 hidden units in the feed-forward layers. 
Training data were collected from the OPUS repository,\footnote{\url{http://opus.nlpl.eu}} and  cleaned with the ModernMT framework \cite{bertoldi2018modernmt}.
At the end of this process, the actual training data is reduced to 45M of segment pairs (550M of English words)
for English-Italian. For English-Spanish, the training data is further filtered with data selection methods \cite{axelrod-etal-2011-domain} using a general-domain seed resulting in 19M segment pairs (330M English words).
Finally, for English-French we have 28M  sentence pairs (550M of English words).
Models are optimized on label-smoothed cross entropy \cite{szegedy2016rethinking} with Adam, with a learning rate that linearly increases for 8,000 updates up to 0.0005, after which decays with inverse square root policy.
Each batch is composed of 4 mini-batches made of 3072 tokens.
Dropout is set to 0.3.
We train for 200,000 updates and average the last 10 checkpoints.
Source and target languages share a BPE~\cite{sennrich-etal-2016-neural} vocabulary of 32k sub-words.

\begin{table*}[hbt!]
\small
\begin{tabular}{l|r|r|r|r|r|r}
 &
  \multicolumn{2}{c|}{\textbf{en-es}} &
  \multicolumn{2}{c|}{\textbf{en-fr}} &
  \multicolumn{2}{c}{\textbf{en-it}} \\
 &
  \textbf{en} &
  \textbf{es} &
  \textbf{en} &
  \textbf{fr} &
  \textbf{en} &
  \textbf{it} \\
  \hline
CARDINAL &
  91 (105) &
  85 (104) &
  87 (101) &
  90 (105) &
  86 (100) &
  85 (98) \\
DATE &
  149 (314) &
  152 (321) &
  145 (303) &
  144 (377) &
  141 (300) &
  141 (294) \\
EVENT &
  8 (26) &
  9 (27) &
  7 (22) &
  7 (27) &
  8 (26) &
  9 (31) \\
FAC &
  18 (31) &
  19 (38) &
  18 (31) &
  21 (52) &
  18 (31) &
  16 (33) \\
GPE &
  241 (338) &
  240 (361) &
  232 (322) &
  221 (312) &
  222 (316) &
  209 (300) \\
LANGUAGE &
  2 (2) &
  2 (2) &
  2 (2) &
  2 (2) &
  2 (2) &
  2 (2) \\
LAW &
  146 (478) &
  141 (622) &
  136 (448) &
  143 (608) &
  137 (439) &
  128 (509) \\
LOC &
  96 (121) &
  91 (122) &
  92 (118) &
  86 (128) &
  89 (111) &
  83 (109) \\
MONEY &
  10 (34) &
  11 (45) &
  10 (34) &
  11 (49) &
  6 (20) &
  6 (25) \\
NORP &
  135 (151) &
  126 (147) &
  136 (151) &
  156 (182) &
  123 (139) &
  143 (194) \\
ORDINAL &
  64 (64) &
  65 (65) &
  57 (57) &
  40 (40) &
  62 (62) &
  52 (53) \\
ORG &
  565 (857) &
  582 (989) &
  550 (844) &
  533 (906) &
  520 (773) &
  485 (851) \\
PERCENT &
  4 (10) &
  4 (6) &
  3 (8) &
  3 (9) &
  4 (10) &
  4 (14) \\
PERSON &
  92 (134) &
  96 (122) &
  88 (129) &
  88 (122) &
  89 (130) &
  87 (101) \\
PRODUCT &
  1 (1) &
  1 (1) &
  1 (1) &
  1 (1) &
  1 (1) &
  1 (1) \\
QUANTITY &
  3 (7) &
  3 (7) &
  3 (7) &
  3 (7) &
  3 (7) &
  3 (7) \\
TIME &
  11 (26) &
  10 (20) &
  10 (22) &
  9 (18) &
  11 (26) &
  11 (24) \\
WORK\_OF\_ART &
  1 (4) &
  1 (4) &
  1 (4) &
  1 (4) &
  1 (4) &
  1 (3) \\
TERM &
  2571 (3174) &
  2662 (3294) &
  2797 (3502) &
  2947 (3659) &
  2166 (2669) &
  2202 (2645) 
\end{tabular}
\caption{\label{tab:stats_ne}
Number of named entities and terms annotated in the test sets (and corresponding number of tokens).}
\end{table*}

\begin{table*}[ht!]
\setlength{\tabcolsep}{3.5pt}
\centering
\small
\begin{tabular}{l|cccc|cccc|cccc}
 & \multicolumn{4}{c}{\textbf{en-es}} & \multicolumn{4}{c}{\textbf{en-fr}} & \multicolumn{4}{c}{\textbf{en-it}} \\
\textbf{} &
  \multicolumn{1}{c}{\textbf{ASR}} &
  \textbf{MT} &
  \textbf{Casc.} &
  \textbf{Dir.} &
  \multicolumn{1}{c}{\textbf{ASR}} &
  \textbf{MT} &
  \textbf{Casc.} &
  \textbf{Dir.} &
  \multicolumn{1}{c}{\textbf{ASR}} &
  \textbf{MT} &
  \textbf{Casc.} &
  \textbf{Dir.} \\
  \hline
CARDINAL      & 92.31  & 88.24 & 80.00 & 76.47 & 94.25  & 86.67 & 80.00 & 81.11 & 93.02  & 89.41 & 78.82 & 80.00 \\
DATE          & 90.60  & 78.95 & 73.68 & 72.37 & 89.66  & 57.64 & 52.08 & 56.25 & 89.36  & 76.60 & 67.38 & 68.09 \\
EVENT         & 37.50  & 33.33 & 33.33 & 33.33 & 28.57  & 71.43 & 28.57 & 57.14 & 37.50  & 66.67 & 44.44 & 55.56 \\
FAC           & 77.78  & 73.68 & 63.16 & 57.89 & 77.78  & 57.14 & 52.38 & 47.62 & 77.78  & 75.00 & 62.50 & 50.00 \\
GPE           & 94.61  & 84.17 & 79.17 & 82.50 & 94.40  & 89.19 & 81.98 & 86.88 & 94.62  & 89.00 & 83.25 & 82.30 \\
LANGUAGE      & 100.00  & 100.00 & 100.00 & 100.00 & 100.00  & 100.00 & 100.00 & 50.00 & 100.00  & 100.00 & 100.00 & 50.00 \\
LAW           & 69.86  & 63.12 & 46.10 & 42.55 & 70.59  & 65.73 & 46.15 & 43.36 & 69.34  & 69.53 & 53.13 & 47.66 \\
LOCATION           & 93.75  & 85.71 & 79.12 & 81.32 & 92.39  & 81.40 & 77.91 & 74.42 & 93.26  & 79.52 & 79.52 & 74.70 \\
MONEY         & 20.00  & 54.55 & 27.27 & 72.73 & 20.00  & 18.18 & 27.27 & 27.27 & 16.67  & 66.67 & 16.67 & 66.67 \\
NORP          & 87.41  & 79.37 & 70.63 & 69.84 & 87.50  & 69.43 & 63.06 & 62.18 & 86.99  & 70.63 & 60.84 & 56.64 \\
ORDINAL       & 90.63  & 81.54 & 72.31 & 69.23 & 89.47  & 80.00 & 65.00 & 70.00 & 90.32  & 80.77 & 65.38 & 65.38 \\
ORG           & 89.38  & 89.00 & 77.49 & 78.69 & 89.09  & 84.08 & 73.97 & 73.36 & 89.23  & 79.59 & 67.63 & 71.96 \\
PERCENT       & 0.00  & 100.00 & 0.00 & 75.00 & 0.00  & 66.67 & 0.00 & 66.67 & 0.00  & 25.00 & 0.00 & 75.00 \\
PERSON        & 40.22  & 93.75 & 40.63 & 38.54 & 39.77  & 93.18 & 38.64 & 38.64 & 39.33  & 98.85 & 42.53 & 41.38 \\
PRODUCT       & 0.00  & 100.00 & 0.00 & 0.00 & 0.00  & 100.00 & 0.00 & 0.00 & 0.00  & 100.00 & 0.00 & 0.00 \\
QUANTITY      & 0.00  & 66.67 & 0.00 & 0.00 & 0.00  & 100.00 & 33.33 & 33.33 & 0.00  & 66.67 & 0.00 & 33.33 \\
TIME          & 63.64  & 100.00 & 80.00 & 70.00 & 60.00  & 77.78 & 77.78 & 66.67 & 63.64 & 63.64 & 63.64 & 45.45 \\
WORK\_OF\_ART & 0.00  & 100.00 & 0.00 & 0.00 & 0.00  & 100.00 & 0.00 & 0.00 & 0.00  & 0.00 & 0.00 & 0.00 
\end{tabular}

\caption{Case insensitive accuracy scores for all the NE types on the three language pairs. We report the results for ASR, MT, Cascade (Casc.) and Direct (Dir.) systems.}
\label{tab:ne_types_results}
\end{table*}

\subsection{Direct ST Model}

Our  \textbf{direct} model has the same architecture of the ASR component described above, which is also used to initialise its encoder weights \cite{bansal-etal-2019-pre}. In addition, we exploit data augmentation  and knowledge transfer techniques  successfully applied by participants in the IWSLT-2020 evaluation campaign \cite{ansari-etal-2020-findings,potapczyk-przybysz-2020-srpols,gaido-etal-2020-end}. For data augmentation, we use SpecAugment and  time stretch~\cite{nguyen2019improving},  together with synthetically generated data  obtained by translating with our NMT model the transcripts contained in the ASR training corpora. Besides encoder pre-training, for knowledge transfer we also apply knowledge distillation (KD): as in \cite{liu2019endtoend,gaido-et-al-2020-on-knowledge}, our \textit{student} ST model is trained by computing the KL divergence \cite{kullback1951} with the output probability distribution of the NMT model used as \textit{teacher}. The whole training procedure is carried out in three phases: starting from the synthetically generated data (with the KD loss function), continuing with  MuST-C and Europarl-ST (still with KD), and concluding with fine-tuning on the same ST data, but  switching to the label-smoothed cross entropy loss.

\section{Statistics for the Annotated Test Sets}
\label{sec:appendix}

Table \ref{tab:stats_ne} presents the number of named entities (NEs) and terms annotated in the test sets, divided by category. 
Since both NEs and terms can be composed of more than one word (e.g. for a person it is common to have both the name and surname),   the total number of tokens per category is also given.

\section{Results for each NE Type}

Table \ref{tab:ne_types_results} shows the accuracy scores for all NE types on all language directions. First, we can notice that some types show a high variability (e.g. LANGUAGE, PRODUCT, QUANTITY), which is caused by the limited number of examples with that label. Otherwise, the performance of the two ST systems (cascade and direct) are similar on all categories, demonstrating that their different architecture does not bring to a different ability in handling specific type of entities.






\end{document}